\documentclass[runningheads]{llncs}
\usepackage[T1]{fontenc}
\usepackage{microtype}
\usepackage{graphicx}
\usepackage{url}
\usepackage[svgnames]{xcolor} 
\usepackage{hyperref}
\usepackage{pdflscape}
\usepackage{tcolorbox}
\tcbuselibrary{skins, breakable}
\usepackage{listings}
\usepackage[labelfont=bf]{caption}
\usepackage[capitalise,nameinlink]{cleveref}
\usepackage{xcolor}
\usepackage{booktabs}   
\usepackage{array}      
\usepackage{amssymb}    
\usepackage{tabularx}
\usepackage{listings}
\usepackage{subcaption}
\usepackage{xcolor}
\usepackage{color}
\usepackage{hyperref}

\usepackage{tikz}
\usetikzlibrary{positioning,fit,calc,arrows.meta}
\pgfdeclarelayer{background}
\pgfsetlayers{background,main}

\hypersetup{
	pdftoolbar=true,        
	pdfmenubar=true,        
	pdffitwindow=false,     
	pdfstartview={FitH},    
	pdfsubject={},   
	pdfcreator={},   
	pdfproducer={}, 
	pdfkeywords={}, 
	pdfnewwindow=true,      
	colorlinks=true,       
	linkcolor=Brown,          
	citecolor=Brown,        
	filecolor=Brown,      
	urlcolor=Brown           
}

\newcommand{\mypar}[1]{\vspace{0.5pt}\noindent\textbf{#1.}}
\newcommand{\mypartwo}[1]{\vspace{0.5pt}\noindent\textit{#1.}}

\usepackage{xcolor}

\usepackage{booktabs}
\usepackage{pifont}

\usepackage{enumitem}
\usepackage{wrapfig}

\begin{document}
\title{Organizational Memory for \\Agentic Business Process Execution}
\titlerunning{Organizational Memory for Agentic Business Process Execution}
%
\author{
Lukas Kirchdorfer\inst{1} \and
Adrian Rebmann \inst{1} \and 
Christian Warmuth \inst{1} \and \\
Timotheus Kampik\inst{1} \and
Theiss Heilker\inst{1} \and
Gregor Berg\inst{1} 
}

\authorrunning{L. Kirchdorfer et al.}

\institute{
    SAP Signavio, Berlin, Germany \\
    \email{\{first.last\}@sap.com} 
    }
\maketitle              
\begin{abstract}
LLM-based agents offer new opportunities for automating business process execution beyond the limits of rule-based systems. However, general-purpose LLMs lack the organization-specific knowledge required for reliable execution, which is typically fragmented across human-oriented artifacts such as policies, process models, and standard operating procedures. While such knowledge can technically be encoded in individual prompts or agent-specific retrieval setups, this approach does not scale in enterprises, as it gives rise to knowledge silos and rule duplicates, and makes consistent updates and learning across agents difficult. We argue that this calls for an \textit{organizational memory} for agentic business process execution: a shared, governed, and agent-consumable reference layer of evolving organization-specific procedural knowledge about how work should be executed. We derive requirements for such a memory, propose an architecture for its curation and consumption, and demonstrate its effectiveness in a proof-of-concept based on a procurement scenario.


\keywords{Agentic Process Automation  \and Large Language Models \and Memory \and Business Process Management.}
\end{abstract}

\section{Introduction}


Organizations are increasingly adopting LLM-based agents to execute business processes~\cite{CALVANESE2026102738,Dumas2026}. 
Unlike deterministic automation approaches such as robotic process automation (RPA), which are confined to structured and repetitive settings, these agents can interpret natural language and autonomously handle multi-step tasks including exception handling. 
Thus, they offer a path to automating processes that were previously difficult to formalize with rule-based systems~\cite{weske2024business}.

However, general-purpose LLMs lack access to organization-specific knowledge by default. 
They are not inherently aware of company policies, process conventions, role structures, system landscapes, or exception-handling practices. 
Yet effective process execution in organizational settings depends precisely on such contextual knowledge, which varies across organizations, departments, and individual teams. 
This knowledge gap poses a fundamental challenge to scaling agentic business process execution in enterprise settings.

A natural response is to encode organization-specific knowledge in prompts or retrieval setups specific to individual agents~\cite{10.1145/3560815,RAG}. 
However, this approach does not scale: in large organizations, thousands of agents are expected to operate simultaneously across evolving processes and heterogeneous knowledge sources. 
Per-agent knowledge encoding creates new silos, where rules are duplicated, updated inconsistently, and interpreted differently across agents. 
At the same time, organizations already possess vast amounts of process-relevant knowledge, represented in BPMN models, standard operating procedures (SOPs), policies, knowledge bases, and process analyses based on historical execution data.
This knowledge is fragmented across artifacts, many of which were designed for human interpretation, not for LLM agent consumption. 
The core challenge is, therefore, not knowledge availability, but its transformation into a unified, agent-consumable form and its consistent provision to all agents in the organization.

We argue that what is needed is an \textit{organizational memory} for agentic business process execution: 
a shared, human-governed, and agent-consumable reference layer of organization-specific procedural knowledge about how work should be executed. 
Such a memory must integrate distributed knowledge from heterogeneous sources into a unified representation that agents can consume at runtime. 
It must provide the right context to the right agent at the right time, rather than flooding agents with irrelevant information. 
Furthermore, it must evolve with organizational change, reflecting policy updates or lessons learned from agent executions. 
Finally, since organizational memory directly shapes agent behavior across processes and departments, its creation and evolution cannot be fully automated and must instead support human governance. 

This paper makes four contributions: (1)~we introduce the concept of \textit{organizational memory for agentic business process execution}, defining it as a unified representation of organizational process knowledge that enables LLM-based agents to execute process tasks in a context-aware manner; (2)~we derive requirements for such an organizational memory; 
(3)~we propose an organizational memory architecture, describing its components, curation, and consumption; and (4)~we demonstrate its effectiveness in a proof-of-concept implementation.

The paper is structured as follows.
\autoref{sec:background} provides background and related work. 
\autoref{sec:motivation} motivates the need for organizational memory, based on which 
\autoref{sec:requirements} defines requirements for an organizational memory architecture, which we present in
\autoref{sec:architecture}.
Then, \autoref{sec:evaluation} showcases the efficacy of our architecture in a proof-of-concept, before
\autoref{sec:conclusion} concludes the paper.

\section{Background and Related Work}
\label{sec:background}
This section provides background and related work on agentic BPM and agentic memory systems.

\subsection{Agentic Business Process Management}
The concept of agents has periodically attracted attention in BPM over the past three decades. Early work studied autonomous software agents that schedule, execute, and monitor business activities~\cite{JenningsFJNOW96}. Later, RPA introduced software robots for automating rule-based tasks~\cite{AalstBH18}. More recently, advances in generative AI and LLMs have renewed interest in agent-oriented process execution.
This development has motivated the emerging paradigm of \emph{Agentic Business Process Management}~\cite{CALVANESE2026102738,VuKLRK25}, which extends traditional BPM by treating autonomous agents as first-class participants in business process execution. 
Thus, rather than viewing processes primarily as predefined control-flow models, agentic BPM emphasizes process execution as the result of autonomous agents that perceive, reason, act, and interact within organizational constraints. 
Several recent studies have advanced this agent-centric perspective through data-driven approaches to process discovery, simulation, and optimization~\cite{TourPKS23,kirchdorferDiscoveringMultiagentSystems2025,KirchdorferDAL26,ShenPLK26a}.
However, these approaches adopt a generic notion of agents rather than specifically instantiating them with LLMs.
In that context, several technical contributions have explored the use of LLMs and LLM-based agents for process automation and execution~\cite{Guan2024,SmartFlow,MontiLMMR24}, yet not considering organization-aware execution. 
More closely related to this work, Kaltenpoth et al.~\cite{Kaltenpoth2025} and Skolik et al.~\cite{SKOLIK2026102748} show that LLM agents can execute business processes more reliably when guided by semi-structured process rules, which they generate from BPMN models or natural-language process descriptions and retrieve from a rule database during execution.
Our work differs in its enterprise-level perspective and the extensiveness of the architecture. While the existing process-rule approaches support process-aware execution for specific agents or processes, our organizational memory treats organization-specific process knowledge as a shared enterprise resource. It integrates many heterogeneous sources, such as policies, SOPs, system documentation, process models, knowledge bases, and execution experience, handles human governance, and addresses lifecycle concerns such as organizational change and reuse across multiple agents and processes.
Moreover, unlike strategy-based guardrailing~\cite{degiacomo2026}, organizational memory does not formally constrain agents; instead, it provides an \textit{informal frame} through additional and useful context.


\subsection{Agentic Memory Systems}
While traditionally, agentic architectures manage symbolic knowledge in \emph{belief bases}~\cite{DBLP:journals/scp/BoissierBHRS13}, recent works tend to focus on \emph{memory} mechanisms for LLM-based agents. Existing approaches manage memory tiers to extend limited context windows~\cite{MEMGPT}, maintain memory streams with dedicated controllers~\cite{SCM}, organize memories dynamically through indexing and linking~\cite{A_MEM}, or store long-term conversational memories to improve personalization and coherence across interactions~\cite{MemoryBank}. 
More broadly, cognitive agent architectures such as CoALA conceptualize memory as a core component of language agents, alongside action spaces and decision-making mechanisms~\cite{CoALA}. 
These works highlight the importance of persistent memory for LLM agents, but primarily focus on individual agents rather than the enterprise-level challenge of curating, governing, maintaining, and consistently providing organization-specific process knowledge across many agents and processes.
Retrieval-augmented generation (RAG)~\cite{RAG} offers a complementary mechanism by grounding LLM outputs in retrieved passages from external document collections. While useful for accessing organizational documents, RAG alone may be insufficient for enterprise process knowledge that is overlapping, evolving, and often relevant beyond the user's immediate query. Different agents may retrieve different chunks, miss cross-cutting rules, or rely on outdated or conflicting sources. Our notion of organizational memory addresses this gap by treating process knowledge as shared, traceable, governed, and reusable memory elements rather than as query-dependent document fragments.




\section{Motivation}
\label{sec:motivation}

In this section, we motivate the need for organizational memory in agentic business process execution by means of a running example, from which we then derive several challenges that arise in this setting.

\begin{figure}[!t]
  \centering
  \includegraphics[width=\linewidth]{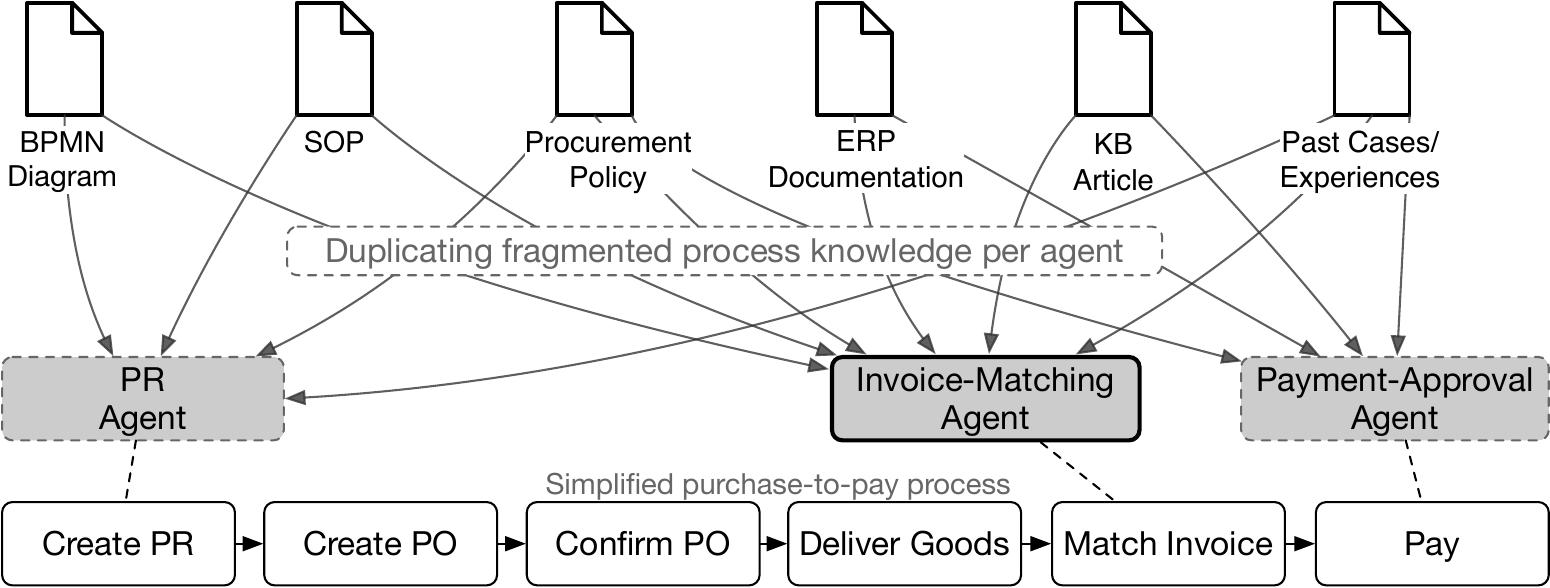}
  \caption{Running purchase-to-pay example: each agent depends on overlapping knowledge that is fragmented across heterogeneous artifacts.}
  \label{fig:motivation}
\end{figure}

\mypar{Problem illustration}
We consider enterprise settings in which LLM-based agents execute business process tasks at runtime, where correct behavior depends on organization-specific procedural knowledge such as policies, responsibilities, systems, and exception-handling practices. Consider a simplified purchase-to-pay process as visualized on the bottom of \autoref{fig:motivation}. 
The process starts with the creation of a purchase requisition (PR), followed by issuing a purchase order (PO) to a vendor. 
The vendor confirms the order, delivers the goods, and submits an invoice. Before a payment can be executed, the invoice must be checked against the corresponding purchase order. 
Suppose an organization wants to automate parts of this process using LLM-based agents. 
One agent may support employees in creating PRs, while another may handle invoice matching.

Focusing on the invoice-matching agent, assume a received invoice does not exactly match the corresponding purchase order. 
The agent must decide whether the invoice can be posted automatically, whether it should be routed to a procurement specialist, or whether it requires compliance review. 
Other agents in the same process, such as a PR agent and a payment-approval agent, depend on overlapping knowledge, as illustrated in \autoref{fig:motivation}. 
At first glance, this appears to be a straightforward reasoning task: compare the invoice to the purchase order, identify deviations, and decide how to proceed. 
However, the correct decision depends on organizational knowledge that is not contained in the LLM itself.

For instance, a BPMN model may specify that invoices with mismatches require an ``invoice clarification'' activity. 
A standard operating procedure may state that price deviations below EUR~250 can be accepted for preferred suppliers. 
A procurement policy may define that deviations involving medical equipment always require compliance review, regardless of their amount. 
ERP documentation may specify where supplier risk classifications are stored, while an internal knowledge-base article may explain that, for a particular subsidiary, the responsible procurement specialist is determined by plant rather than by product category. 
Relevant knowledge may also emerge from previous executions, for example, when human experts repeatedly resolve a recurring mismatch in the same way or when agents receive corrections for exceptional cases. 
Each of these sources contains potentially relevant knowledge, but no single source provides the complete execution context required by the agent.

\medskip
\mypar{Challenges}
Several challenges arise in this setting:

\mypartwo{Fragmented and human-oriented knowledge}
Process-relevant knowledge is distributed across heterogeneous sources. These sources typically differ in format, terminology, granularity, and quality, and they are primarily created for human interpretation rather than for direct consumption by LLM-based agents.
Moreover, such knowledge is often not complete and up-to-date, and sometimes too extensive for humans to review in its entirety.
In the purchase-to-pay example, the agent must combine knowledge from a BPMN model, a procurement policy, an SOP, ERP documentation, subsidiary-specific instructions, and potentially past exception handling to make a single invoice-handling decision. The challenge is therefore to transform fragmented organizational knowledge into a unified, agent-consumable representation.

\mypartwo{Shared knowledge across agents and processes}
Organization-specific knowledge is rarely relevant to only one agent. The same procurement policy, supplier classification, approval rule, or compliance constraint may be needed by multiple agents operating in different parts of the same process or in related processes. If each agent is equipped with its own prompt, local memory, or retrieval setup, organizations risk creating new knowledge silos: rules are duplicated, updated inconsistently, and interpreted differently across agents.
In the purchase-to-pay example, a PR agent, an invoice-matching agent, and a payment approval agent may all depend on overlapping procurement and finance rules. These rules should therefore be maintained once and made available consistently, rather than being reimplemented separately per agent.

\mypartwo{Organizational knowledge change and learning}
Organizational knowledge is not static. Policies are updated, responsibilities shift, systems change, and new exception-handling practices emerge. This phenomenon is well-known in BPM under \textit{concept drift}~\cite{KRAUS2026102584}. Consequently, the knowledge used by agents must evolve with the organization. For instance, if the company changes the acceptable mismatch threshold from EUR~250 to EUR~100, all affected agents must reliably act according to the updated knowledge.
The challenge is to keep the shared knowledge layer current without requiring each agent to be updated individually.

\mypartwo{Runtime accessibility and efficiency}
Agents require organization-specific knowledge at the moment of execution, but only the knowledge that is relevant to the current task and context. Simply providing all potentially relevant documentation is neither scalable nor reliable, as it increases latency, introduces irrelevant information, and may distract the agent from the actual decision. In the example, the invoice agent should receive the applicable mismatch tolerance, the supplier-specific rule, the relevant compliance constraint, and the responsible organizational role, but not unrelated procurement guidelines or ERP manuals.

\mypartwo{Human governance}
Since organization-specific knowledge affects how agents behave across processes and departments, its creation and evolution cannot be left entirely to automated mechanisms. Human decision-makers must be involved in deciding which knowledge should be added, modified, or discarded, and how conflicts between sources should be resolved. This raises the question of who is allowed to access, approve, and change which parts of the memory. In the purchase-to-pay example, procurement experts may be responsible for invoice-matching rules, finance for payment policies, compliance for regulatory constraints, and local process owners for subsidiary-specific exceptions. At the same time, not every user who can read a rule should also be able to modify it, and changes made by one organizational unit may affect agents operating in another. Organizational memory therefore requires governance mechanisms for ownership, access control, approval workflows, and accountability.

\medskip
\noindent
The challenges show that reliable and scalable agentic process execution requires more than agent-specific prompts or local memories.
It calls for a shared organizational memory of procedural knowledge about how work should be executed.

\section{Architecture Requirements}
\label{sec:requirements}

Based on the challenges described above, we derive the requirements for an organizational memory architecture.

\mypartwo{R1: Integration of heterogeneous organizational knowledge}
Integrate process-relevant knowledge from heterogeneous sources. In the example, this includes the BPMN model, procurement policy, invoice-matching SOP, ERP documentation, and past exception handling.


\mypartwo{R2: Agent-consumable representation}
Transform fragmented, often human-oriented process knowledge into a unified representation that is decomposed into self-contained units of procedural knowledge, so LLM-based agents can interpret and apply it during process execution. 
Unification alone (e.g., turning sources to plain text) is insufficient, since without decomposition into self-contained units, individual rules cannot be addressed, compared, updated, or selectively retrieved. 
For instance, a BPMN model and a textual rule should not only be expressed in a common format but also be broken down to the level of individual rules. 

\mypartwo{R3: Conflict detection and consistency management}
Support the identification and resolution of inconsistencies, redundancies, and conflicts between memory fragments 
originating from different sources, which is a general problem in AI~\cite{CoreaKTG24}. For example, if an SOP allows deviations below EUR~250 while a plant-specific policy defines a lower threshold, this conflict must be detected and resolved.

\mypartwo{R4: Traceability to source artifacts}
Preserve links between memory fragments and their original sources to enable transparency. In the purchase-to-pay example, an agent's decision to route an invoice to compliance should be traceable to the underlying policy document.

\mypartwo{R5: Context-specific retrieval}
Provide agents with the organization-specific knowledge relevant to the current process task and context. For instance, the invoice-matching agent should receive the applicable mismatch tolerance, supplier-specific rule, compliance constraint, and responsible role, but not unrelated HR guidelines.

\mypartwo{R6: Low-latency runtime accessibility}
Provide organizational memory efficiently at runtime so agents can access relevant context without prohibitive latency.

\mypartwo{R7: Support for adaptive memory}
Support the continuous evolution of organizational memory over time, both through deliberate updates to policies, organizational structures, and systems, and through insights gained from process execution experiences. For example, if the allowed mismatch threshold changes from EUR~250 to EUR~100, all affected agents should follow the updated rule after a single change to the memory. Similarly, repeated human corrections for a recurring invoice mismatch may indicate a newly observed exception or successful practice and become a candidate update to the organizational memory.



\mypartwo{R8: Human governance and ownership}
Support human oversight of the creation and modification of organizational memory. This includes defining ownership of memory fragments, specifying read and write permissions, supporting approval workflows for updates, and ensuring accountability for changes. For instance, procurement department may own invoice-matching rules, whereas finance department may own payment policies.

\mypartwo{R9: Scalability across agents and processes}
Scale to enterprise settings in which multiple agents operate across different processes, departments, organizational units, and information systems while relying on a shared organizational memory. 
In the example, a PR agent, invoice-matching agent, and payment approval agent should all be able to access the same maintained procurement and finance rules.

\section{Organizational Memory Architecture}
\label{sec:architecture}
\begin{figure}[t!]
    \centering
    \includegraphics[width=\linewidth, trim={1cm 0.5cm 1cm 0.5cm}]{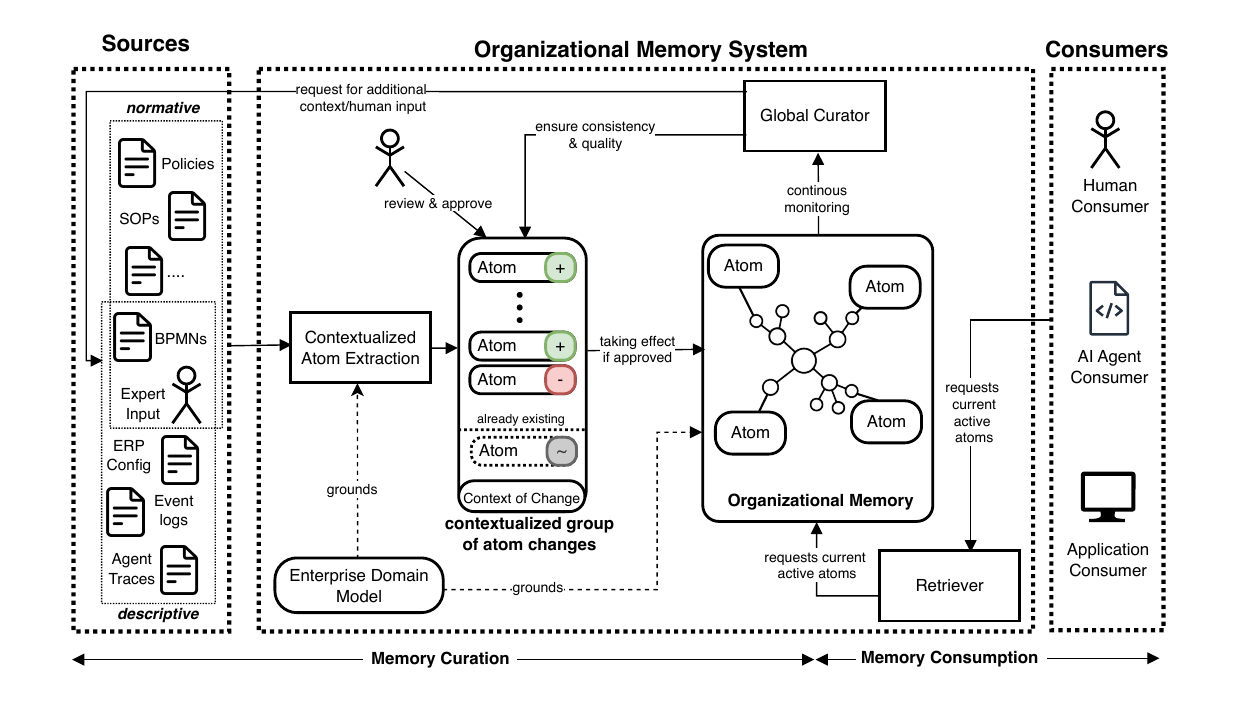}
    \caption{
    Organizational memory architecture.
    }
    \label{fig:arch}
\end{figure}

This section introduces the proposed organizational memory architecture, addressing the requirements outlined above. \autoref{fig:arch} provides a high-level overview of the architecture. 
It shows how heterogeneous sources are transformed into fragments, which we call \textit{process atoms} (\textit{atoms} for short), that collectively constitute the organizational memory. 
This memory can then be accessed by consumers, such as LLM-based agents, to support organization-specific process execution.
In the following, we first make the core design choices underlying the architecture explicit in~\autoref{sec:rationale}, before describing how organizational memory is curated in~\autoref{sec:curation} and how it is consumed during process execution in~\autoref{sec:consumption}.

\subsection{Design Choices}
\label{sec:rationale}

Before describing the architecture in detail, we make core design choices explicit, justifying them against the requirements.

\mypar{Atomic decomposition}
Memory fragments could in principle be represented as plain text, document chunks, or larger pieces of structured process descriptions. 
We instead represent each fragment as a \textit{process atom}: a self-contained unit that captures exactly one rule, constraint, or responsibility, with its applicability conditions separated from its required action and purpose. 
Concretely, each process atom has the following core attributes:

\begin{itemize}[noitemsep, topsep=0pt]
    \item \textit{Name}: a short name for the desired process behavior;
    \item \textit{Source}: a reference to the input source from which the atom was extracted;
    \item \textit{Content}: a structured textual description with three parts:
    \begin{itemize}[noitemsep, topsep=0pt, label=$-$]
        \item \textit{Applicability}: the context in which the atom applies (e.g., process, activity, role, organizational unit, business object, or precondition);
        \item \textit{Action}: the concrete rule, constraint, or required behavior;
        \item \textit{Purpose}: the rationale for the rule, i.e., why it exists.
    \end{itemize}
\end{itemize}

\noindent
Atomic decomposition addresses R2 and enables the requirements that operate at the level of single rules and behavioral constraints: 
conflict detection between candidate and existing knowledge (R3), 
targeted updates as policies change (R7), 
per-rule traceability to source artifacts (R4), 
and retrieval of the smallest sufficient set for a given task (R5). 
A plain-text representation can be unified and agent-consumable, but it makes it difficult to be addressed, compared, or updated at the rule level without decomposition. 
Conversely, decomposing further than the described level, would lose the business meaning that domain experts rely on to govern memory (R8), even if a finer logical decomposition would be possible. 
This idea of representing procedural knowledge as atomic units is, among others, also followed in declarative process modeling and analysis~\cite{DBLP:books/sp/22/CiccioM22}; we provide an integrated structure, i.e., name, source, content, and tags (described below), tailored to organizational memory for LLM-based agents.

\mypar{Domain-model grounding via tags}
Atoms must be grounded in organizational concepts so they can be retrieved in a specific context, for the right process, role, business object, or unit. 
We ground atoms through categorized \textit{tags} that jointly represent an \textit{Enterprise Domain Model}. 
A natural alternative is to ground atoms in a formal ontology or a knowledge graph with explicit relations~\cite{Bein2026}. 
A formal ontology requires upfront schema design and continuous ontology engineering, though, which is a strong requirement. A tag vocabulary can be assembled incrementally and easily revised as the organization changes (R7). Tags also align well with the way LLM-based retrievers and agents consume context. 

\mypar{Retrieval pattern}
At runtime, we follow a (hybrid) retrieval pattern rather than a specific mechanism. We combine filtering using the domain-model tags with semantic or LLM-based selection within the filtered set. 
Semantic search over atom content in isolation does not scale well and cannot reliably distinguish, e.g., an invoice-matching rule that applies in subsidiary~A from one that applies in subsidiary~B, because that distinction may be encoded in the organizational metadata, not the rule text. 
Pure tag-based filtering cannot resolve fine-grained relevance among many atoms that share the same tags. 
The architecture therefore features a \textit{Retriever} that takes a context request and returns a relevant atom set, leaving the concrete mechanism to instantiation.\footnote{Note that we do not claim or evaluate that a single mechanism is best in this paper.}

\subsection{Memory Curation}
\label{sec:curation}

This section describes how organizational memory is created and maintained. As shown in~\autoref{fig:arch}, the curation phase takes multiple input sources, extracts atoms, creates a group of atom changes, and finally updates the memory through a human-governed review and change process.

\mypar{Input sources}
The curation phase can take different types of organizational knowledge as input, including policies, SOPs, process models, system documentation, knowledge-base articles, and traces from previous process executions. These sources differ in format, structure, terminology, and degree of formality. 

\mypar{Step 1: Create group of atom changes}
In the first step, we create a contextualized group of atom changes by deriving a set of candidate process atoms from the input sources and contextualizing these with the existing atoms in the organizational memory. Since different source types encode knowledge differently, source-specific extraction services are required. For instance, a BPMN model must be analyzed differently from a policy document or an execution trace. Note that the design of extraction services for all possible input sources is beyond the scope of this paper. 
For simplicity, we focus on PDF documents, such as policies and SOPs, which also serve as the input sources in our proof-of-concept evaluation in~\autoref{sec:evaluation}. 
In general, the purpose of this step is to identify process-relevant rules, constraints, responsibilities, and exception-handling knowledge and represent them as atoms (see~\autoref{sec:rationale} for the atom representation).



\noindent
Assume a finance control directive states that every PR must by assigned a valid cost center. The extraction service may create the following atom:

\begin{itemize}[noitemsep, topsep=0pt]
    \item \textit{Name}: Ensure that a cost center is assigned on each PR.
    \item \textit{Source}: Finance Control Directive, Section 3.
    \item \textit{Content}:
    \begin{itemize}[noitemsep, topsep=0pt, label=$-$]
        \item \textit{Applicability}: All purchase requisitions, in all company codes.
        \item \textit{Action}: Every PR must be assigned to a valid cost center before it can be submitted; if the requester does not provide a cost center, the PR must be rejected.
        \item \textit{Purpose}: Ensures that every expenditure is traceable to a responsible organizational unit for budgetary control and audit purposes.
    \end{itemize}
\end{itemize}

\noindent
The extraction services also ground each candidate in the \textit{Enterprise Domain Model} through one or more categorized tags, such as \texttt{procurement} (corporate function), \texttt{purchase requisition} (business object), or \texttt{purchase-to-pay} (end-to-end process). These tags provide a shared semantic vocabulary for organizing and later retrieving atoms from the memory. 

Candidate atoms are then assessed by the \textit{Global Curator}, an active entity in our architecture that has read access to the existing organizational memory. The Global Curator performs two complementary checks. First, it carries out quality checks on each candidate atom in isolation, such as verifying that the atom has a meaningful name, a clearly defined scope, and a single rule. Second, it relates each candidate to the existing memory, checking whether it duplicates, overlaps with, or conflicts with atoms already stored.
For example, if a candidate atom, which is extracted from a policy document, is already present in the organizational memory, it does not need to be added again. 
Conversely, if a candidate atom conflicts with an existing atom, the conflict must be detected and resolved, for instance by updating the existing atom, replacing it, or flagging the conflict for human review. 
The outcome of this step is a proposed set of atom changes to the memory, which may include atoms to be added, removed, or modified.


\mypar{Step 2: Update organizational memory}
Given the contextualized group of atom changes, a \textit{Human} expert reviews the changes, checks the atoms and their sources, and decides if changes should be applied. They may accept the proposal, modify individual atoms, reject individual changes, or resolve conflicts differently than suggested by the curator. 
In the purchase-to-pay example, a finance expert may approve changes to cost-center rules, while procurement or compliance experts may be responsible for invoice-matching or regulatory constraints. 
This human-in-the-loop design ensures that the memory remains governed and that changes to the shared knowledge layer are overseen by experts.

\mypar{Organizational memory}
The output of the curation phase is the updated organizational memory: a governed collection of validated atoms grounded in the enterprise domain model. Through their tags and source references, atoms can be organized, traced, and retrieved for runtime use. 
The memory thereby acts as the shared knowledge layer from which agents can obtain task-specific organizational context during process execution.
The \textit{Global Curator} can continuously monitor the organizational memory and request additional context or human input.

\subsection{Memory Consumption}
\label{sec:consumption}
This section describes how organizational memory is consumed during process execution. As shown in~\autoref{fig:arch}, memory consumption starts with a context request from the runtime agent, continues with a query to the organizational memory, and results in a set of relevant atoms that is provided to the agent as execution context.

\mypar{Step 1: Create context request}
At runtime, an agent is assigned to execute parts of a business process. 
To do so reliably, it may require organization-specific knowledge that is not contained in its prompt or model knowledge. The agent therefore creates a context request for the \textit{Retriever}. 
This request describes the current execution context, including the process, activity, agent role, organizational unit, input data, and case state. 
In the purchase-to-pay example, an invoice-matching agent may include information about the supplier, invoice amount, purchase order, material category, subsidiary, and current activity.

\mypar{Step 2: Query organizational memory}
The \textit{Retriever} uses the context 
request to identify atoms in the \textit{Organizational Memory} that are applicable to the current situation. Different retrieval mechanisms may be used for this purpose, such as tag-based filtering, semantic search, rule-based matching, or LLM-based selection. The tags assigned during memory curation are particularly useful because they ground atoms in the enterprise domain model and make them retrievable by process, activity, role, object type, or organizational context. The goal is to retrieve the smallest sufficient set of atoms for the current task, rather than exposing the agent to all potentially related documentation. Reliably avoiding both under-retrieval and over-retrieval is a key challenge for enterprise-level settings and motivates future research.

\mypar{Step 3: Provide set of atoms}
The \textit{Retriever} returns the selected atoms to the runtime agent. These atoms are appended to the agent's context and guide its decision-making and activity execution.

In summary, the proposed architecture addresses the requirements froms~\autoref{sec:requirements} through its two complementary phases. The curation phase integrates heterogeneous input sources (R1), transforms them into unified atoms (R2), checks candidate atoms against the existing memory for duplicates and conflicts (R3), preserves source references (R4), and supports governed updates as organizational knowledge changes (R7--R8). The runtime consumption phase enables context-specific retrieval (R5) with low overhead (R6) by providing agents with a compact set of relevant atoms rather than full documents. Finally, the shared organizational memory addresses scalability across agents and processes (R9), since multiple agents can rely on the same maintained knowledge layer instead of separate local memories or retrieval setups.



\section{Proof-of-Concept}
\label{sec:evaluation}

This section presents a preliminary proof-of-concept of whether the proposed organizational memory architecture can improve the ability of LLM-based agents to execute business process tasks in an organization-aware manner.\footnote{The source code for our implementation cannot be made publicly available due to proprietary reasons. Yet, we provide a detailed description of the artifact’s design and the experimental procedure in the paper, and supply additional evaluation details (incl. data and ground truth): \url{https://github.com/lukaskirchdorfer/org-memory-paper}} 
Specifically, we examine whether access to organizational memory supports agents in making decisions that are in line with desired behavior during process execution.
We describe the experimental setup in~\autoref{sec:exp_setup} and the results in~\autoref{sec:results}.

\subsection{Experimental Setup}
\label{sec:exp_setup}
We use the purchase-to-pay process introduced in~\autoref{sec:motivation}, where an LLM-based agent acts as a procurement assistant supporting employees in creating purchase requisitions (PRs) in the company's ERP system. 
Given a natural-language request from a user, the agent must decide whether the requested purchase complies with the organization's rules. If so, the agent should create the corresponding PR with the correct information. If the request violates organizational rules, the agent should refuse the creation and explain the reason for non-compliance.

\mypar{Evaluation data}
To create a controlled evaluation environment, we construct a synthetic document collection representing the organizational knowledge base of a hypothetical company. 
The collection consists of 9 manually created PDF documents, comprising 5 procurement-related documents and 4 distractor documents from other organizational functions, such as HR. The procurement-related documents resemble typical enterprise artifacts, such as policies and SOPs, and encode 11 distinct rules governing the purchase-to-pay process. These rules cover approval thresholds, supplier restrictions, category-specific requirements, and exception-handling procedures.
The rules were manually specified and embedded in realistic document contexts rather than provided as isolated rule lists. This design allows us to evaluate whether the proposed architecture can identify and retrieve relevant organizational knowledge from unstructured enterprise documents. The distractor documents are included to reflect a broader enterprise knowledge base and to assess whether the architecture can provide task-relevant context without exposing the agent to excessive or irrelevant information. 

\mypar{Scenarios}
We designed 10 scenarios, each representing a specific employee request to the agent. 
The scenarios cover varying levels of complexity: they range from straightforward compliant requests, in which the employee provides all required information, to requests that violate a single policy, and finally to cases in which the agent must jointly consider multiple policies to determine the correct response.
Each scenario is annotated with a binary expected outcome---\textsc{Create} or \textsc{Refuse}---and a corresponding ground-truth specification of PR field values (material, quantity, vendor) that makes evaluation possible in a rule-based manner. 
The ground truth was created manually by the authors based on the synthetic policy documents, ensuring that each decision and annotated field value can be traced directly to a specific clause in the source documents.



\mypar{Agent setups}
We compare three setups (\textit{Base}, \textit{RAG}, and \textit{Memory}) using the same LLM and system prompt, yet, differing in organizational knowledge access: 

\mypartwo{Base: No access to organizational process knowledge}
The agent receives no policy information and must decide whether to create a PR based solely on the information in the employee request and the LLM's parametric knowledge.

\mypartwo{RAG: RAG-based access to organizational process knowledge}
Before each request, the agent retrieves the top-5 most relevant chunks from a vector database that contains the document corpus via dense retrieval embeddings.
The ChromaDB vector database has been created in a preprocessing step from the given input documents. Each document is split into overlapping chunks of approximately 600 characters using a paragraph-aware chunker with a 100-character overlap between adjacent chunks, and all chunks are embedded with \texttt{text-embedding-3-large} using cosine similarity as the distance metric.
  
\mypartwo{Memory: Access to Organizational Memory (ours)}
Before each request, the agent requests the \textit{Retriever} to provide the set of necessary atoms for the specific procurement situation. The atoms have been automatically extracted and created from the given input documents.

Note that although granting the agent direct access to all documents is a conceivable setup, we deliberately exclude it from our evaluation because it is neither realistic nor scalable in enterprise settings.




\mypar{LLMs}
We instantiate all three agents with both OpenAI's GPT-4.1 and Anthropic's Claude Sonnet~4.5 and report results for both models.

\mypar{Metrics}
We evaluate agent behavior using the \textit{Policy Compliance Rate} (PCR), defined as the fraction of scenarios in which the agent produces the policy-compliant outcome. For \textsc{Create} scenarios, the agent is considered correct only if it creates a PR with the expected material, quantity, and vendor, which we verify through rule-based comparison against ground-truth fields. For \textsc{Refuse} scenarios, the agent is considered correct only if it refrains from creating a PR.


\subsection{Results}
\label{sec:results}

\begin{wrapfigure}{r}{0.5\textwidth}
    \centering
    \includegraphics[width=\linewidth]{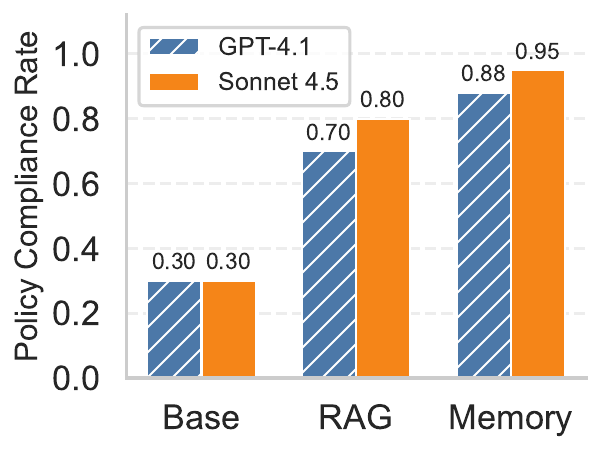}
    \caption{Results across all 10 scenarios.}
    \label{fig:results}
\end{wrapfigure}

\autoref{fig:results} reports the PCR for all three agent configurations, averaged over ten scenarios and four runs per scenario. The organizational memory agent achieves the highest PCR for both models, reaching 88\% with GPT-4.1 and 95\% with the more advanced Sonnet~4.5. It clearly outperforms both the 
base agent, which reaches 30\% for both models, and the RAG-based agent, which reaches 70\% with GPT-4.1 and 80\% with Sonnet~4.5. This indicates that structured organizational memory can improve policy-compliant process execution beyond both parametric model knowledge and document-level retrieval.
The base agent performs poorly, often creating PRs even when the request violates a policy. Its correct decisions mostly occur in simple scenarios where the relevant information is explicit in the employee request and no organization-specific reasoning is required. 
RAG improves substantially over the base setup, showing that document retrieval can help when the relevant policy is semantically close to the employee request. However, RAG still fails in scenarios where the required rule is cross-cutting rather than query-specific. For example, when an employee requests notebooks 
without specifying a cost center, the RAG agent creates the PR, although the organization requires a cost center for every PR. Since this universal rule is stated in a finance policy and is not semantically close to the notebook request, it is not retrieved.

The organizational memory agent mitigates this limitation by retrieving task-relevant atoms rather than relying solely on similarity between the request and document chunks. This allows it to consider both request-specific rules and generally applicable constraints, such as the mandatory cost center requirement. The memory agent also handles scenarios that require multiple policies to be considered jointly. Its remaining errors are mainly over-restrictions: in some runs, the agent refuses a request that is actually compliant. Inspection shows that this is caused by an imprecisely extracted atom whose applicability scope is too broad. This highlights that the effectiveness of organizational memory depends not only on runtime retrieval, but also on the quality of memory curation.

Overall, the results provide initial evidence that organizational memory can improve organizational-knowledge-aware process execution, while at the same time,
it is not designed to and cannot ensure correct process execution in all cases.
Notably, the results should be interpreted with caution, given the preliminary nature and limited scope of the evaluation.

\section{Conclusion}
\label{sec:conclusion}

LLM-based agents create new opportunities for automating business process execution, but reliable execution in enterprise settings requires access to organization-specific procedural knowledge. In this paper, we introduced \textit{organizational memory} as a shared, governed, and agent-consumable reference layer for such knowledge. We derived requirements for organizational memory, proposed an architecture that covers both memory curation and runtime consumption, and instantiated the architecture through process atoms that can be extracted from organizational artifacts, governed by humans, and retrieved by agents during execution. A proof-of-concept in a procurement scenario indicates that organizational memory can improve policy-compliant agent behavior.

Our work is an initial step and should be interpreted in light of its limitations. The proof-of-concept covers a single synthetic procurement process with a small policy corpus consisting only of PDF documents. Thus, generalization to real enterprise settings with numerous overlapping policies, process models, conflicting sources, and legacy documentation is not yet demonstrated. Future work should therefore evaluate organizational memory in larger and more realistic enterprise environments.
Beyond larger-scale evaluation, our work opens several directions for future research at both the engineering and management levels. From an engineering perspective, our experiments indicate that execution quality strongly depends on the precision of atom extraction and curation. Future research should therefore develop robust extraction mechanisms for a broader range of organizational knowledge sources beyond PDF documents. From a management perspective, future work should investigate how organizational memory can be governed effectively and efficiently by human experts in large organizations, including questions of responsibility, accountability, lifecycle management, update propagation, and scalability as organizational knowledge evolves over time. 


%
%
%
\bibliographystyle{splncs04}
\bibliography{bibliography}

\end{document}